\documentclass[letterpaper, 10 pt, conference]{ieeeconf}
\IEEEoverridecommandlockouts
\usepackage{graphicx} 
\usepackage{amsmath,amssymb}
\usepackage{cite}
\usepackage{booktabs}
\usepackage[ruled, vlined, linesnumbered]{algorithm2e}
\usepackage{multirow}
\usepackage{multicol}
\usepackage{xcolor}
\usepackage{pifont}   
\usepackage{amsfonts}
\usepackage{algorithmic}
\usepackage{array}
\usepackage{siunitx}
\usepackage{textcomp}
\usepackage{stfloats}
\usepackage{url}
\usepackage{verbatim}
\usepackage{graphicx}
\usepackage{amsfonts,bm}
\usepackage{tikz}
\usepackage{color}
\usepackage{mathtools}
\usepackage{rotating}
\usepackage{blkarray}
\usetikzlibrary{arrows}
\usepackage{makecell}

\usepackage{amsthm}
\usepackage{comment}

\usepackage{subcaption} 

\graphicspath{{images/}}

\title{\LARGE \bf
TransTac: Visuo-Tactile Modality Transition via Ultraviolet-Encoded Transparent Elastomers
}

\author{
Lingyue Yang$^{1}$ and Bin Fang$^{1}$%
\thanks{This work was supported by the Brain Science and Brain-like Intelligence Technology - National Science and Technology Major Project (Grant No. 2025ZD0215600), the National Natural Science Foundation of China under Grant No.62573063, 62536001, the Open Foundation of the State Key Laboratory of Precision Space-time Information Sensing Technology No.STSL2025-B-07-01, and the Open Projects Program of State Key Laboratory of Multimodal Artificial Intelligence Systems.}%
\thanks{$^{1}$Beijing University of Posts and Telecommunications, Beijing, China.
Email: yly-valentina@bupt.edu.cn, fangbin1120@bupt.edu.cn}%
\thanks{Corresponding author: Bin Fang (fangbin1120@bupt.edu.cn).}%
}

\begin{document}

\maketitle
\thispagestyle{empty}
\pagestyle{empty}

\begin{abstract}

Vision-based tactile sensors (VBTS) recover high-resolution contact geometry but typically rely on opaque elastomer layers that prevent visual transparency, while RGB-D cameras provide global depth perception yet degrade significantly at close range. To address this limitation, we present \textit{TransTac}, a transparent ultraviolet (UV)-encoded binocular VBTS that integrates visual observation and marker-based tactile reconstruction within a single compact device. The system employs a transparent elastomer embedded with UV-reflective markers and a prior-guided Delaunay stereo matching algorithm for robust sparse triangulation.

To reliably detect densely distributed semitransparent markers, we develop a lightweight detector that enables stable localization under contact and deformation. The proposed prior-guided Delaunay matching improves correspondence robustness by approximately 21\% compared with global assignment baselines while maintaining high reconstruction accuracy. In semantic evaluation, TransTac achieves up to 83.3\% zero-shot recognition accuracy on tactile images, exceeding opaque tactile baselines by approximately 50 percentage points. Embedding analysis further reveals substantially stronger cross-modal alignment with natural images, with class-center similarity increasing from around 0.2 to over 0.77. Controlled near-distance experiments quantify the degradation of RGB-D depth reliability and demonstrate extended geometric coverage enabled by visuo-tactile integration. Finally, a compact prototype is implemented with an approximate hardware cost of \$70. 

Code and hardware design are publicly available at https://github.com/87361/TransTac.

\end{abstract}
\section{Introduction}

Tactile sensing is essential for robotic manipulation, providing information about object shape \cite{li2024jacquard}, contact state \cite{donlon2018gelslim}, and material properties \cite{zheng2024materobot,zhang2023tirgel} that vision alone cannot reliably capture. Vision-based perception often degrades under occlusion \cite{zhang2017learning} or challenging illumination, preventing reliable estimation of contact geometry during manipulation. While tactile feedback can compensate for such failures, most existing robotic systems still lack compact and multimodal tactile sensors capable of delivering both accurate contact geometry and appearance-related cues at the interaction interface. This limitation restricts robust grasping and adaptive policy selection in unstructured environments.

Existing vision-based tactile sensors (VBTS) are largely geometry-oriented. 
Systems such as GelSight \cite{yuan2017gelsight} and its variants \cite{lambeta2020digit} reconstruct high-resolution contact deformation using coated elastomers, enabling tasks such as slip detection \cite{james2018slip} and fine-grained shape reconstruction \cite{lu2025stereotactip,do2022densetact}. 
However, the opaque reflective coating blocks visual observation through the sensing surface, requiring separate vision sensors for appearance and global geometry estimation.

\begin{figure}
    \centering
    \includegraphics[width=1\linewidth]{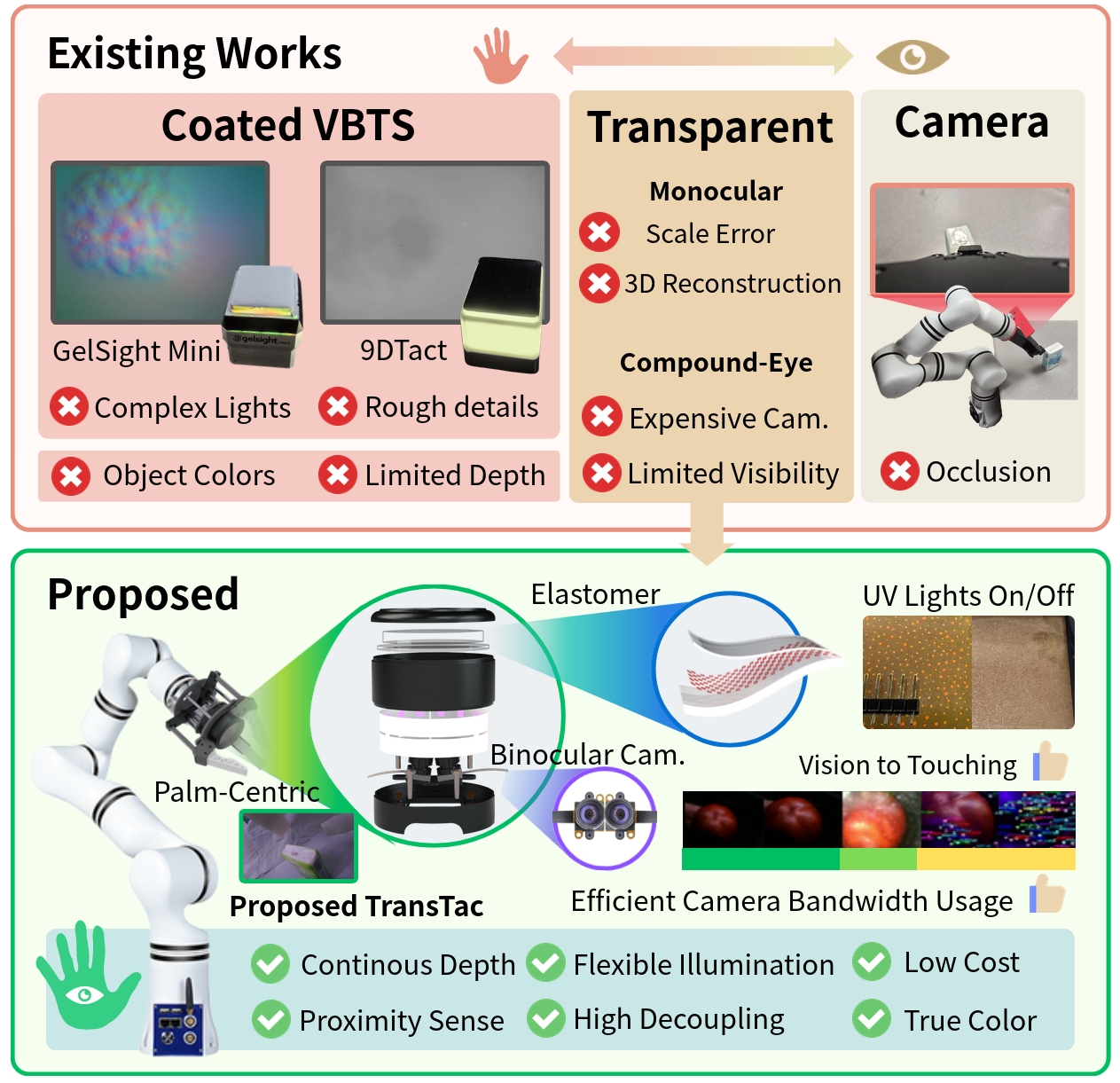}
    \caption{
    Motivation of \textbf{TransTac}. RGB-D cameras provide global depth but degrade at close range, while coated VBTS reconstruct only contact deformation. TransTac integrates binocular vision and UV-encoded transparent tactile sensing within a unified hardware structure.
    }
    \label{fig:Motivation}
\end{figure}

The core \textbf{challenge} lies in maintaining reliable geometric perception across the transition from pre-contact observation to physical contact. RGB-D cameras provide global depth estimation but experience reduced depth reliability at close range, particularly near the minimum sensing distance. In contrast, coated VBTS capture local deformation only after contact and cannot recover recessed or non-contact geometry. This separation between far-field depth estimation and near-field tactile reconstruction creates a geometric sensing gap.

As illustrated in Fig.~\ref{fig:Motivation}, existing RGB-D perception and VBTS designs each address only part of this range. Bridging this gap requires a sensing configuration that preserves visual transparency, enables robust marker-based geometric tracking, and supports metric depth recovery in near-contact scenarios.

In this work, we present \textbf{TransTac}, a transparent ultraviolet-encoded VBTS that integrates binocular RGB observation and marker-based tactile reconstruction within a compact stereo configuration. The transparent elastomer preserves visual clarity, while embedded UV-reflective markers enable robust geometric tracking under switchable illumination. We introduce a Delaunay-based stereo marker matching algorithm to establish reliable correspondences under epipolar constraints, enabling sparse metric triangulation of the contact surface. By combining real-time marker displacement tracking with stereo-based depth estimation and RGB-D fusion, TransTac supports both dynamic tactile interaction monitoring and near-contact geometric reconstruction.

\noindent\textbf{Our contributions are summarized as follows:}
\begin{itemize}
    \item \textbf{A transparent ultraviolet-encoded binocular VBTS hardware design.} 
    We present a stereo vision-based tactile sensor built upon a transparent elastomer embedded with UV-reflective markers, enabling simultaneous visual observation and marker-based tactile reconstruction within a single compact hardware structure.

    \item \textbf{A lightweight detection model for dense semitransparent markers.} 
    We develop a compact marker detection network tailored for densely distributed semitransparent markers, achieving robust localization under varying illumination and contact conditions.

    \item \textbf{A prior-guided Delaunay stereo matching algorithm.} 
    We introduce a Delaunay-based marker correspondence method initialized by epipolar nearest-neighbor matching, enabling reliable sparse triangulation and metric contact-surface reconstruction under geometric constraints.
\end{itemize}

\section{Related Work}

\subsection{Vision-Based Tactile Sensing}

VBTS recovers contact geometry by observing elastomer deformation with cameras. Representative systems such as GelSight \cite{yuan2017gelsight} and the GelSlim series \cite{donlon2018gelslim,taylor2022gelslim} employ reflective coatings and photometric stereo to reconstruct high-resolution surface geometry. Learning-based mappings have further been introduced to infer contact shape from calibrated visual signals, as demonstrated by DIGIT \cite{lambeta2020digit} and the DTact family \cite{lin2022dtact,lin20239dtact}. While these designs achieve accurate local deformation sensing, they typically rely on opaque reflective layers and controlled illumination.

Transparent or vision-through tactile sensors have also been explored. Yamaguchi and Atkeson \cite{yamaguchi2016combining} demonstrated a transparent optical tactile skin capable of both deformation sensing and external visual observation. However, maintaining reliable geometric reconstruction while preserving visual transparency remains challenging, especially in near-contact scenarios where visual and tactile signals may interfere.

\subsection{Marker-Based Tactile Reconstruction}

Embedding markers within deformable membranes provides an alternative approach to tactile geometry estimation. Systems such as FingerVision \cite{sun2019fingervision}, TacTip \cite{ward2018tactip}, and TAC2Pose \cite{lee2023tac2pose} track marker displacement to infer contact information, while transparent implementations such as CompdVision \cite{luo2024compdvision} explore marker-based reconstruction under optical constraints.

Reliable marker detection under deformation and illumination variation remains challenging. Dense marker patterns may undergo geometric distortion during contact, affecting localization accuracy. Prior work has explored robust detection mechanisms \cite{li2023real} and illumination-based separation strategies such as UV-reflective markers in SpecTac \cite{wang2022spectac}. Nevertheless, consistent detection and correspondence of dense semitransparent markers under switchable illumination remain difficult.

\subsection{Stereo Correspondence and Near-Contact Depth Sensing}

Reliable stereo correspondence is essential for triangulating marker positions in vision-based tactile sensing systems. Under rectified stereo geometry, correspondence search can be restricted to one-dimensional scanlines along epipolar lines. Classical approaches include global assignment methods such as Hungarian matching and support-point propagation strategies such as ELAS \cite{geiger2010efficient}. However, when applied to marker-based tactile sensing, correspondence estimation becomes more challenging due to repetitive marker appearance, dense spatial distribution, and deformation-induced geometric distortion.

In parallel, RGB-D sensing provides dense depth estimation but suffers from fundamental limitations in near-contact scenarios. Because of triangulation constraints and minimum sensing distance, the reliability of depth measurements degrades rapidly as objects approach the camera \cite{servi2021metrological}. As a result, RGB-D sensors often fail to provide stable geometric cues in the immediate pre-contact region.

These limitations motivate combining tactile triangulation with visual depth estimation to achieve continuous geometric perception from pre-contact observation to physical contact.

\section{Design and Fabrication}

\begin{figure*}
    \centering
    \includegraphics[width=1\linewidth]{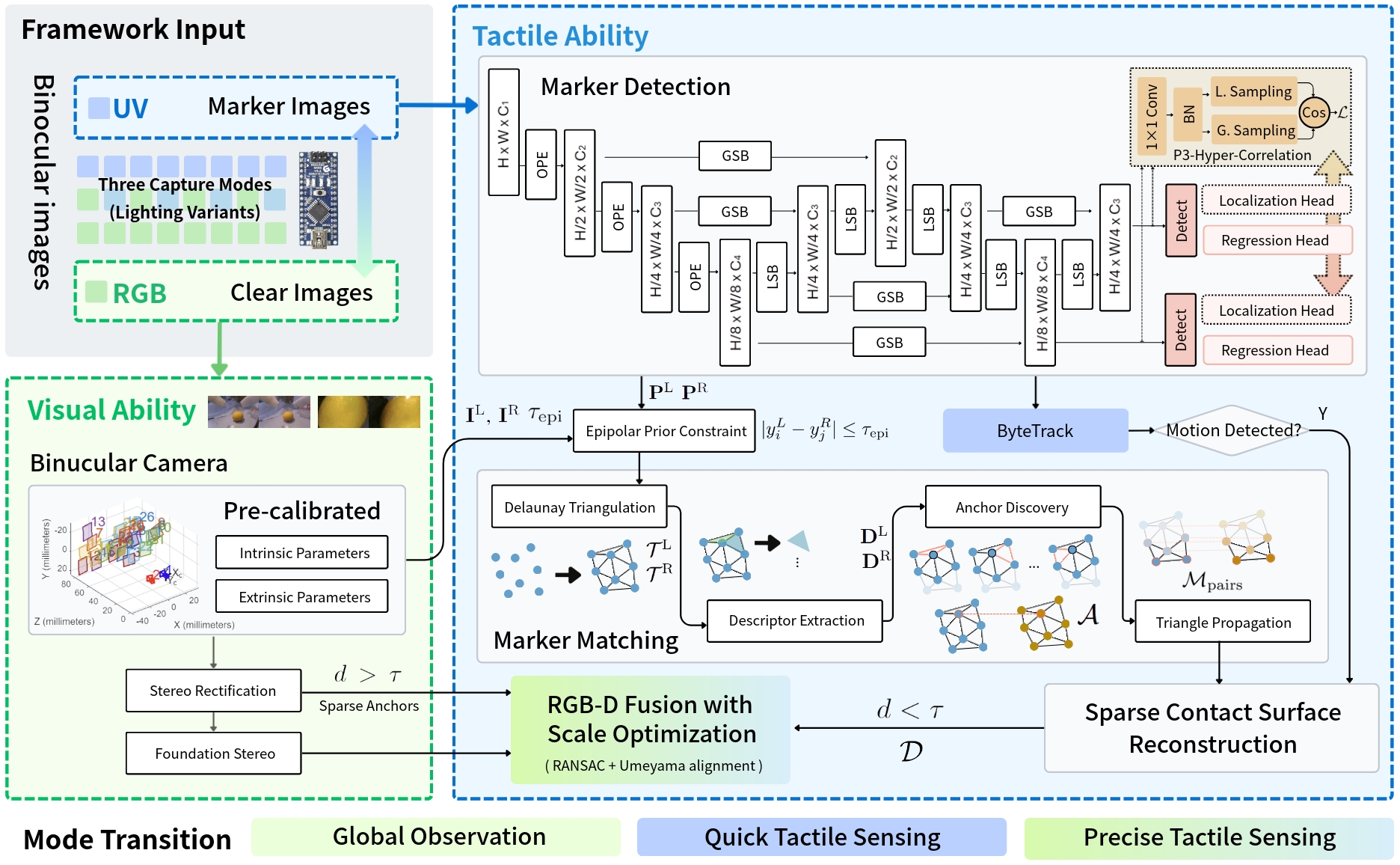}
    \caption{Overview of the TransTac framework. Binocular RGB and UV images are rectified for marker detection and stereo correspondence. Sparse triangulated depth is fused with RGB-D estimation for contact-surface reconstruction. Color-coded modules indicate the final processing modules in each sensing stage.}
    \label{fig:flowchart}
\end{figure*}

\subsection{TransTac Design Principles}

TransTac is designed to integrate visual observation and tactile reconstruction within a compact, low-cost stereo configuration. A transparent silicone membrane embedded with UV-reflective markers preserves optical clarity while enabling marker-based deformation sensing. Time-multiplexed illumination separates RGB imaging from UV marker observation without requiring additional optical components.

As illustrated in Fig.~\ref{fig:flowchart}, binocular images are first rectified and processed for marker localization and stereo correspondence. Marker displacement provides local deformation cues, while triangulated correspondences enable sparse metric depth estimation. These components jointly support contact-surface reconstruction within a unified sensing interface.

\subsection{Sensor Hardware and Fabrication}

The sensing unit comprises a stereo pair of USB camera modules (HBVCAM-2436-R V11) with a horizontal field of view of $100^\circ$ and manual focus. Intrinsic and extrinsic parameters are obtained through standard multi-view calibration using a checkerboard of size GP052.

The tactile interface is fabricated from transparent silicone elastomer (Solaris\textsuperscript{™}, 1:1 ratio) cast into a 3D-printed mold with a thickness of 5~mm. Fluorescent markers are manually embedded at irregular spatial locations and encapsulated within a thin silicone layer to reduce light attenuation. Illumination is provided by a 365~nm ultraviolet strip and a white LED strip.

The outer housing is 3D-printed in matte black material to suppress internal reflections and boundary artifacts.

\subsection{Marker Detection and Tracking}

Accurate localization of densely distributed semitransparent markers is critical for reliable stereo correspondence and deformation tracking. Compared with conventional blob-based methods, the proposed setting presents additional challenges, including marker translucency, UV/RGB illumination switching, and deformation-induced appearance variation.

\paragraph{Network architecture}
As illustrated in Fig.~\ref{fig:flowchart}, we adopt a lightweight single-stage, anchor-free detector built upon a compact convolutional backbone \cite{zhou2025nnwnet}. A correlation modeling module is introduced to capture both local and higher-order spatial relations among densely packed small markers. The detection head follows a CenterNet-style design \cite{duan2019centernet}, consisting of a Gaussian-encoded center heatmap branch and a regression branch for bounding box refinement. Skip connections are employed to preserve fine-grained spatial details.

\paragraph{Training objective}
The overall loss is defined as
\begin{equation}
\mathcal{L} = 
w_{\text{loc}} \mathcal{L}_{\text{loc}} 
+ w_{\text{reg}} \mathcal{L}_{\text{reg}} 
+ w_{\text{corr}} \mathcal{L}_{\text{corr}},
\end{equation}
where the localization term supervises Gaussian-encoded center prediction, the regression term refines bounding box geometry, and the correlation term enforces feature consistency among markers within the same local cluster. This formulation improves robustness under deformation and partial occlusion without significantly increasing computational cost.

\paragraph{Tracking}
Detected markers are associated across frames using ByteTrack\cite{zhang2022bytetrack} to maintain identity consistency and to estimate displacement for tactile deformation analysis.

\subsection{Prior-Guided Stereo Marker Matching}
\label{subsec:marker_matching}

Given rectified stereo images $\mathbf{I}^L$ and $\mathbf{I}^R$, let
$\mathcal{P}^L = \{\mathbf{p}^L_i\}$ and
$\mathcal{P}^R = \{\mathbf{p}^R_j\}$
denote the detected marker centroids in the left and right views.
The objective is to establish reliable stereo correspondences
$\mathcal{M} = \{(\mathbf{p}^L_i,\mathbf{p}^R_j)\}$ under epipolar and geometric constraints.

\paragraph{Epipolar Prior Initialization}

For each marker $\mathbf{p}^L_i$, candidate matches in the right image
are first restricted by the epipolar constraint

\[
|y^L_i - y^R_j| \le \tau_{\mathrm{epi}} .
\]

Among valid candidates, the nearest neighbor in disparity space
is selected to form an initial correspondence set $\mathcal{M}_0$.
These high-confidence matches provide statistical priors for
subsequent anchor discovery.

\paragraph{Delaunay Encoding and Anchor Discovery}

Delaunay triangulation is applied to $\mathcal{P}^L$ and $\mathcal{P}^R$,
producing triangle sets $\mathcal{T}^L$ and $\mathcal{T}^R$.
For a triangle $t_{ijk}$ we define a normalized edge-length descriptor

\[
\mathbf{d}_{ijk} =
\frac{
[\|\mathbf{p}_i-\mathbf{p}_j\|,\;
 \|\mathbf{p}_j-\mathbf{p}_k\|,\;
 \|\mathbf{p}_k-\mathbf{p}_i\|]
}{
\|[\|\mathbf{p}_i-\mathbf{p}_j\|,
  \|\mathbf{p}_j-\mathbf{p}_k\|,
  \|\mathbf{p}_k-\mathbf{p}_i\|]\|_2
}.
\]

Triangle pairs satisfying descriptor similarity and epipolar
consistency are selected as anchor seeds $\mathcal{A}$.

\paragraph{Triangle Propagation}

Starting from anchor pairs, correspondences are propagated
to adjacent triangles while enforcing geometric consistency.
This local structural propagation preserves neighborhood
topology under moderate deformation and avoids ambiguity
in global assignment.

The resulting set $\mathcal{M}$ provides robust stereo correspondences for
subsequent sparse triangulation and dense depth alignment.

\subsection{RGB-D Fusion with Scale Optimization}

The sparse triangulated marker points provide reliable geometric depth near the contact region, while dense depth is estimated from binocular RGB images using the pretrained stereo model FoundationStereo~\cite{wen2025foundationstereo}. However, the predicted dense depth may suffer from scale bias.
To combine the advantages of both sources, we align the dense depth map with geometrically reliable sparse triangulated depths.

\paragraph{Distance-Aware Sparse Anchors}
The switching threshold $\tau$ is defined as the calibrated
distance between the stereo cameras and the elastomer surface.
For $d>\tau$, visual stereo features provide sparse anchors,
while for $d<\tau$ triangulated markers with detectable
displacement (tracked by ByteTrack) are used instead.

\paragraph{Sparse Triangulation}
Given matched marker pairs $(p_i^L,p_i^R)$ and calibrated stereo projection matrices

\[
P_L = K_L[I|0],\quad P_R = K_R[R|t],
\]

each correspondence is triangulated to obtain a sparse 3D point

\[
X_i^{\text{sparse}}=(X_i,Y_i,Z_i).
\]

These triangulated points provide metrically reliable depth observations near the contact region.

\paragraph{Dense Depth Alignment}

Meanwhile the RGB-D model predicts a dense depth map $D_{net}(u,v)$. Each pixel is back-projected into a dense 3D point

\[
X_{dense}(u,v)=
\left(
\frac{u-c_x}{f_x}Z,
\frac{v-c_y}{f_y}Z,
Z
\right),\quad Z=D_{net}(u,v).
\]

For each triangulated feature location $(u_i,v_i)$ we obtain the corresponding dense point $X_i^{dense}$, producing sparse–dense correspondences $(X_i^{sparse},X_i^{dense})$.

After removing outliers using RANSAC, a similarity transformation

\[
T=(s,R,t)
\]

is estimated via Umeyama alignment to minimize

\[
\sum_i \|X_i^{sparse}-(sRX_i^{dense}+t)\|^2.
\]

The transformation is then applied to all dense points to obtain a metrically consistent dense depth map.

\section{Experiments and Results}

We evaluate the proposed TransTac system across four key capabilities:
(i) semantic recognizability of tactile imagery,
(ii) recovery of near-contact geometry in regions where RGB-D sensing fails,
(iii) robustness of stereo marker correspondence estimation, and
(iv) stability of tactile marker tracking during interaction.

Experiments are designed to analyze both perceptual and geometric aspects of the proposed visuo–tactile sensing framework. 
Semantic evaluation focuses on whether tactile observations preserve high-level visual semantics that can be interpreted by modern vision-language models. 
Geometric evaluation investigates the system’s ability to recover accurate contact geometry in near-contact scenarios where conventional RGB-D sensors exhibit depth degradation.

For experiments involving geometric reconstruction, ground-truth geometry is obtained from STL models of 3D-printed objects. 
This provides sub-millimeter reference accuracy without relying on noisy active depth measurements.

For quantitative experiments, the sensor is rigidly mounted on a three-axis gantry system that provides repeatable positioning with sub-millimeter precision along the Z-axis. 
This setup enables controlled evaluation of depth estimation and reconstruction stability at varying distances. 
For qualitative experiments, the sensor is handheld to allow diverse contact poses, sliding interactions, and realistic manipulation scenarios.

\subsection{Semantic Recognizability of Tactile Images}

We first evaluate whether tactile images captured by TransTac preserve sufficient visual semantics for object-level recognition. 
Two vision-language models (Qwen-VLM and ChatGPT-VLM) and two open-vocabulary detectors (YOLO-World\cite{cheng2024yolo} and YOLO-E\cite{wang2025yoloe}) are used to assess semantic interpretability across tactile modalities.

Three sensing modalities are considered: GelSight Mini, 9DTact, and the proposed TransTac sensor. 
Six object categories (egg, coin, battery, Lego block, button, and glass bead) are used, with two instances per category. 
Each object is captured from three viewpoints, producing 36 tactile images for each modality.

Since tactile images may contain partial object cues, recognition performance is evaluated using a \emph{weighted semantic scoring scheme}. 
Predictions that partially match the ground-truth category receive proportional credit, allowing fair comparison across sensing modalities and models.

Table~\ref{tab:semantic_results} summarizes both recognition results and embedding-level alignment. 
Across all recognition models, TransTac consistently achieves substantially higher semantic scores than opaque tactile sensors. 
For example, ChatGPT-VLM achieves 83.3\% on TransTac images, compared with 30.2\% for GelSight and 12.5\% for 9DTact.

To further analyze representation-level consistency, we compute embedding similarity using SigLIP2\cite{tschannen2025siglip} and DINOv2\cite{oquab2023dinov2}. 
TransTac tactile images remain significantly closer to natural image representations, with DINOv2 class-center similarity increasing from approximately 0.20--0.24 for opaque tactile sensors to 0.774 for TransTac. 
Nearest-neighbor alignment also reaches 100\%, indicating strong cross-modal consistency between tactile and visual representations.

\begin{table}[t]
\caption{Semantic preservability across tactile modalities.
All recognition models are evaluated using a weighted semantic scoring scheme.}
\centering
\small
\setlength{\tabcolsep}{4pt}
\renewcommand{\arraystretch}{1.05}

\begin{tabular}{lccc}
\toprule
\textbf{Model} & \textbf{GelSight} & \textbf{9DTact} & \textbf{TransTac} \\
\midrule

\textit{Recognition} & & & \\
Qwen-VLM      & 28.7 & 10.8 & \textbf{80.6} \\
ChatGPT-VLM   & 30.2 & 12.5 & \textbf{83.3} \\
YOLO-World    & 15.3 & 2.1  & \textbf{72.2} \\
YOLO-E        & 16.9 & 3.6  & \textbf{75.0} \\

\midrule

\textit{Embedding} & & & \\
SigLIP2 (Zero-shot \%) & 27.8 & 27.8 & \textbf{83.3} \\
DINOv2 (Center Similarity) & 0.236 & 0.202 & \textbf{0.774} \\
DINOv2 (NN Top-1 \%) & 27.8 & 55.6 & \textbf{100.0} \\

\bottomrule
\end{tabular}
\label{tab:semantic_results}
\end{table}

\begin{figure}[t]
    \centering
    \begin{subfigure}[t]{\linewidth}
        \centering
        \includegraphics[width=\linewidth]{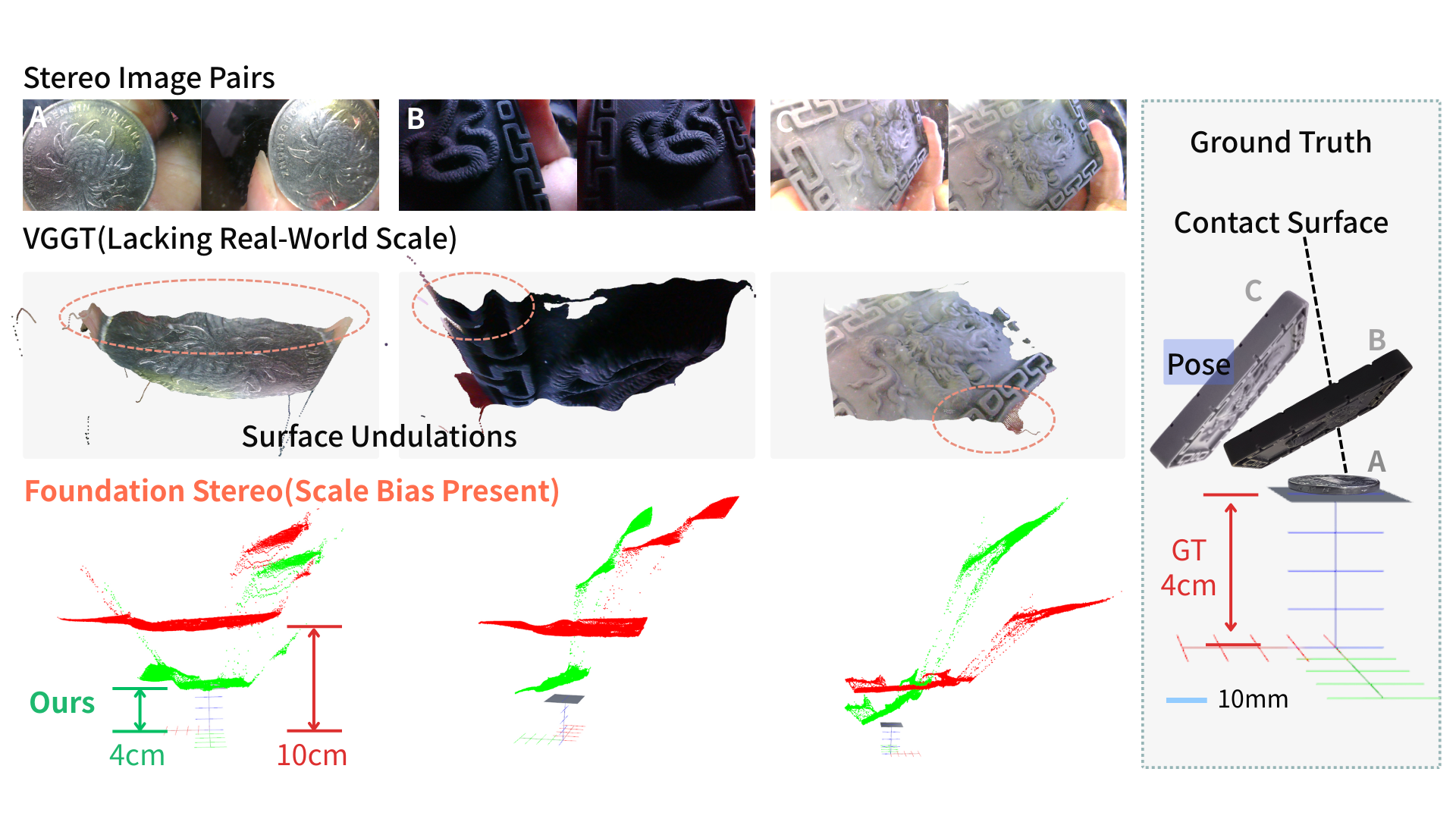}
        \caption{Scale ambiguity in learning-based methods.}
    \end{subfigure}
    \begin{subfigure}[t]{\linewidth}
        \centering
        \includegraphics[width=\linewidth]{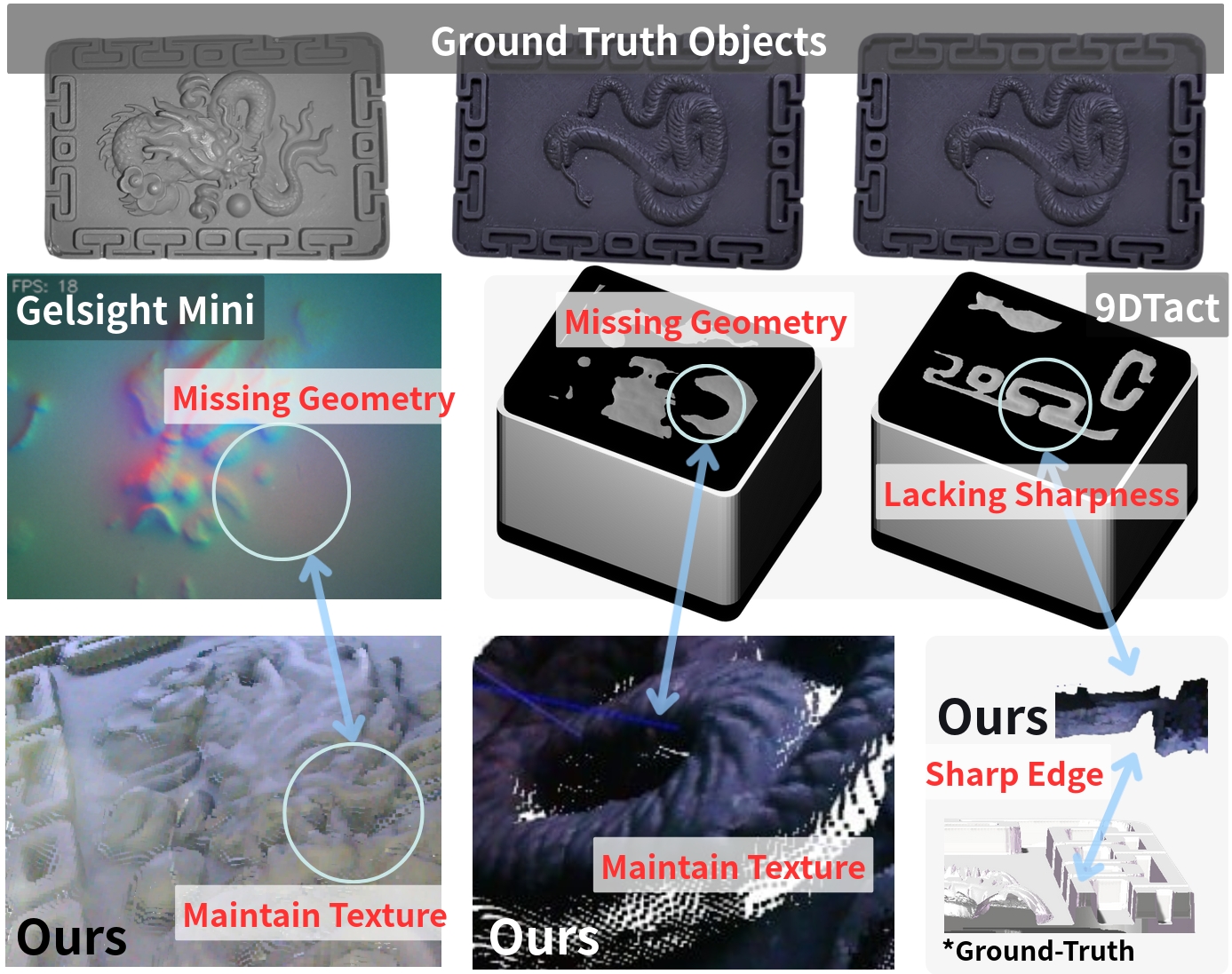}
        \caption{Limitations of coated VBTS in perceiving recessed geometry.}
    \end{subfigure}
    \caption{Qualitative comparison of contact surface reconstruction.}
    \label{fig:reconstruction_comparison}
\end{figure}

\subsection{Near-Contact Geometry Recovery}

Most RGB-D sensors (e.g., Intel RealSense and Microsoft Kinect~\cite{smisek20113d}) have minimum sensing distances that extend beyond the near-contact region. Even sensors capable of closer-range operation exhibit rapid degradation as objects approach the camera, leaving a sensing gap immediately before physical contact.

Coated vision-based tactile sensors (VBTS) suffer from a different limitation. Because the sensing layer is opaque and planar, they can only observe the contact interface and cannot capture recessed surface geometry~\cite{sun2025tactile,lu2025stereotactip}.

To illustrate these limitations, we compare TransTac with both RGB-D depth estimation methods and coated VBTS sensors. Fig.~\ref{fig:reconstruction_comparison}(a) shows that recent depth estimation models such as VGGT~\cite{wang2025vggt} and FoundationStereo~\cite{wen2025foundationstereo} struggle in this setting. VGGT fails to recover the correct metric scale, interpreting perspective cues as planar projections, while FoundationStereo preserves relative structure but exhibits absolute scale errors due to the short stereo baseline and distribution mismatch.

Fig.~\ref{fig:reconstruction_comparison}(b) compares our approach with coated VBTS sensors. These sensors exhibit two typical artifacts: (i) recessed geometries are missing because only protruding regions contact the sensing surface, and (ii) elastic deformation smooths sharp edges into slopes, producing false depths along object boundaries.

To quantify the near-contact limitation of RGB-D sensing, we measure the valid depth ratio of an Intel RealSense D405 as the distance between the sensor and a planar object decreases. The valid depth ratio is defined as the proportion of pixels with non-zero depth values within the region of interest.

\begin{figure}[t]
\centering
\includegraphics[width=\linewidth]{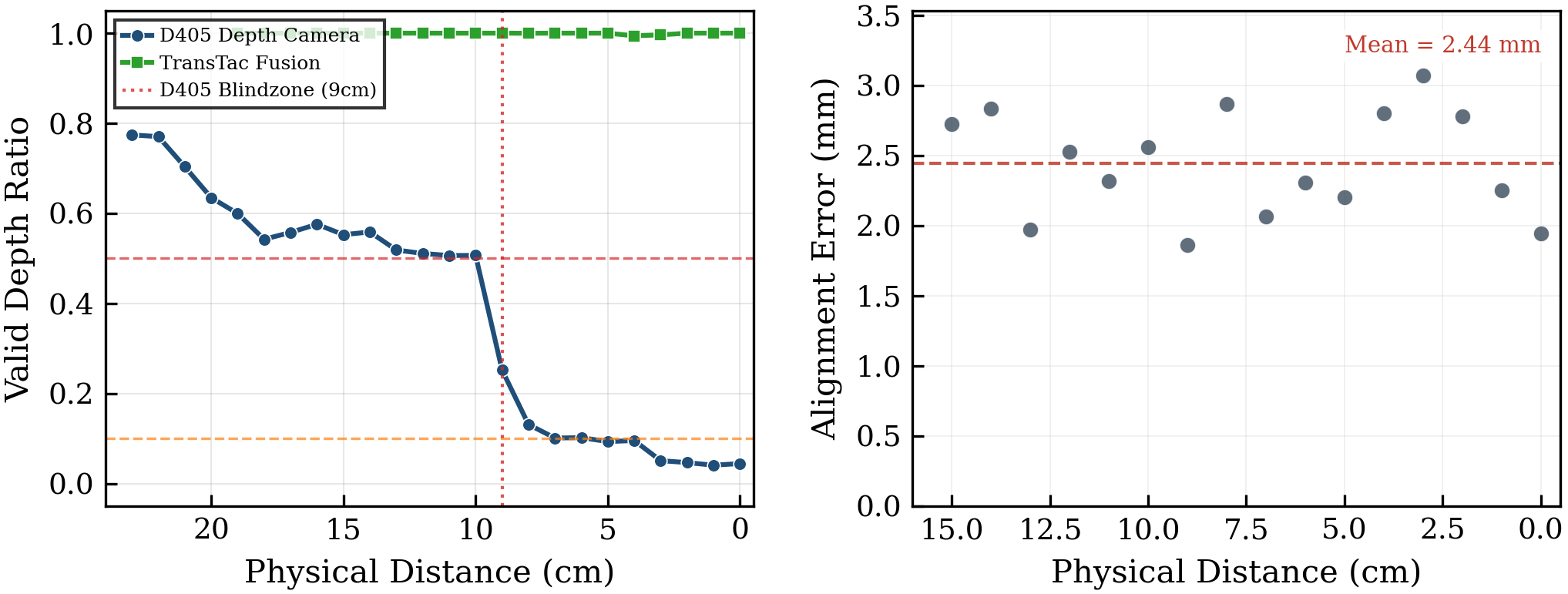}
\caption{
Near-contact depth sensing and alignment evaluation.
(a) Valid depth ratio of an Intel RealSense D405 versus distance.
(b) Sparse–dense alignment error across distances after RANSAC filtering.
}
\label{fig:depth_coverage_alignment}
\end{figure}

Fig.~\ref{fig:depth_coverage_alignment}(a) plots the valid depth ratio as a function of physical distance. Depth coverage remains relatively stable when the object is farther than approximately $9$\,cm from the sensor. As the object enters the near-contact region, however, the proportion of valid depth pixels drops sharply and eventually falls below $10\%$, revealing a near-contact sensing blind zone where RGB-D sensing becomes unreliable.

Despite this degradation, sparse triangulated markers still provide stable geometric constraints. Fig.~\ref{fig:depth_coverage_alignment}(b) shows the alignment error between triangulated marker depths and dense depth predictions after scale alignment. Across all evaluated distances, the mean geometric error remains approximately $2.44\,\mathrm{mm}$, indicating that reliable geometric cues can still be obtained in the near-contact region.

Overall, TransTac combines RGB-D estimation with marker-based triangulation to provide continuous geometric observations from approach to physical contact.

\subsection{Stereo Marker Matching Robustness}

Reliable stereo marker correspondence is critical for accurate triangulation of marker positions. 
However, dense semitransparent markers embedded in a deformable elastomer introduce several challenges for correspondence estimation, including repetitive appearance, dense spatial distribution, and geometric distortion caused by elastomer deformation during contact.

To evaluate the robustness of the proposed matching strategy, we compare the prior-guided Delaunay matching with several baseline approaches. The evaluated baselines include epipolar nearest-neighbor matching and global assignment using the Hungarian algorithm. We also include representative dense correspondence methods, including optical flow (Farneback) and stereo block matching (SGBM), for comparison.

The benchmark is constructed using stereo marker observations collected under diverse interaction conditions, including varying contact pressures, sliding motions, and partial occlusions. Ground-truth correspondences are manually annotated for evaluation.

Since the stereo cameras observe the transparent elastomer layer from slightly different viewpoints, a small portion of markers near the image boundaries may appear in only one view. In addition, the markers are randomly distributed across the sensing surface, so the set of markers visible in the two images is not perfectly identical. As a result, the number of markers that can be matched between the stereo pair varies slightly across frames.

Table~\ref{tab:matching_benchmark} reports the average number of correctly matched markers. The proposed prior-guided Delaunay matching achieves the best performance with 90.8 correct matches on average, substantially improving over Hungarian assignment with 74.9 matches and epipolar nearest neighbor with 74.5 matches.

Dense correspondence methods perform noticeably worse in this scenario. Optical flow and stereo matching obtain 37.9 and 28.7 correct matches respectively. This behavior is mainly caused by the near-contact imaging geometry, where the left and right cameras observe different object surfaces. Regions visible in one view may correspond to different physical surfaces in the other view, reducing the number of reliable dense correspondences.

\begin{table}[t]
\caption{
Stereo marker correspondence comparison under different matching strategies.
Results report the average number of correctly matched markers.
}
\label{tab:matching_benchmark}
\centering
\small
\setlength{\tabcolsep}{4pt}
\renewcommand{\arraystretch}{1.05}
\begin{tabular}{lc}
\toprule
\textbf{Method} & \textbf{Avg. Correct Matches} \\
\midrule
\textit{Marker-Based Matching} \\
\textbf{Prior-Guided Delaunay (Ours)} & \textbf{90.8} \\
Hungarian Assignment & 74.9 \\
Epipolar Nearest Neighbor & 74.5 \\
\midrule
\textit{Dense Correspondence} \\
Optical Flow (Farneback) & 37.9 \\
Stereo Matching (SGBM) & 28.7 \\
\bottomrule
\end{tabular}
\end{table}

	\begin{figure*}
	    \centering
	    \includegraphics[width=1\linewidth]{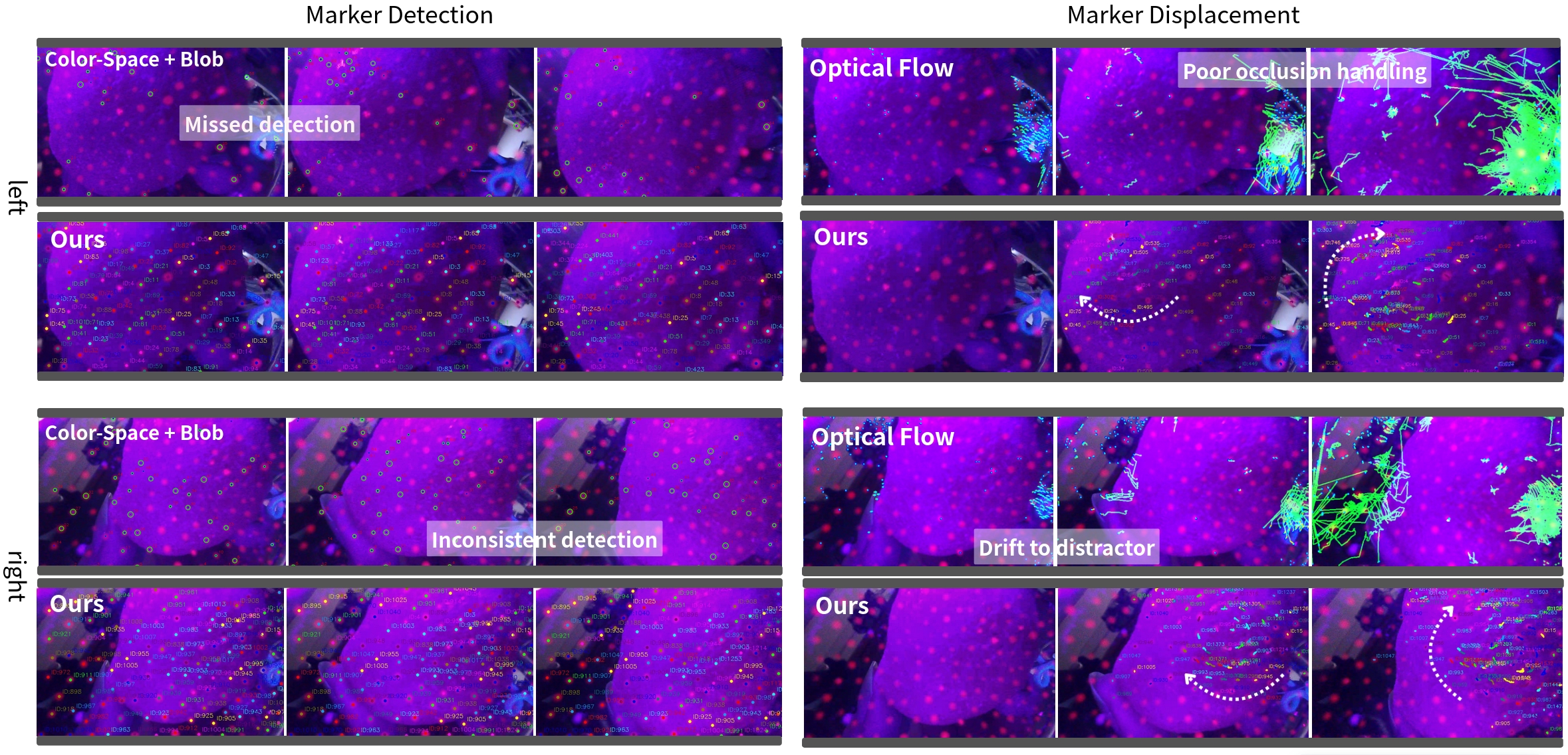}
	    \caption{Qualitative comparison: Left) Marker detection: traditional methods (prone to missed/inconsistent detections) vs our robust model. Right) Trajectory stability under slip/deformation: optical flow (drift/identity swaps) vs our stable tracking.}
	    \label{fig: result}
	\end{figure*}

\subsection{Tactile Marker Tracking}

To assess tactile sensing quality, we track markers embedded beneath the transparent silicone layer of the sensor under diverse interaction conditions including varying contact pressure, tangential slip, rolling motion, and large deformation. The proposed method is compared with a dense optical flow baseline that has been widely used in prior work. The proposed detection model operates at $20$\,FPS in real time on a laptop GPU with NVIDIA RTX 4060 architecture.

We benchmark our marker detection capability against the traditional approach of color space thresholding combined with blob detection, which was employed in prior work utilizing UV-activated markers\cite{wang2022spectac}. This conventional method exhibits several limitations, including frequent missed detections under varying illumination conditions or partial occlusion, as well as inconsistent marker identification and localization across frames.

The optical flow baseline computes pixel displacements between consecutive frames, making it susceptible to cumulative drift and unable to reliably separate tactile-induced motion from background visual changes. In contrast, our method applies displacement estimation, isolating marker movement associated with actual contact deformation. This decoupling prevents false positives from background motion or viewpoint changes. In non-contact cases, occasional specular highlights are filtered out, as they do not correspond to physically adhered contact points.

Qualitative sequences in Fig.~\ref{fig: result} illustrate the difference: our system maintains stable marker identities and trajectories throughout challenging slip and deformation scenarios, whereas the optical flow baseline exhibits rapid drift, identity swaps, and intermittent loss of correspondences.

\section{Conclusion}

We presented TransTac, a transparent ultraviolet-encoded binocular vision-based tactile sensor that integrates marker-based tactile reconstruction and stereo observation within a compact hardware structure. The system combines semitransparent marker detection, prior-guided Delaunay stereo matching, and sparse triangulation with RGB-D fusion to support contact-surface reconstruction and dynamic marker tracking.

Extensive experiments validate the effectiveness of the proposed system. In semantic evaluation, TransTac achieves up to 83.3\% zero-shot recognition accuracy on tactile images across the evaluated models. Embedding-level analysis further confirms semantic preservation. The DINOv2 center similarity increases from around 0.2 for opaque VBTS to 0.774 with TransTac, indicating substantially improved alignment with natural image representations.

In stereo matching benchmarks, the proposed prior-guided Delaunay method improves correspondence robustness by approximately 21\% compared with global assignment baselines while maintaining high matching accuracy. Controlled near-distance measurements quantify the degradation of RGB-D depth validity at close range and further show that the proposed visuo-tactile integration extends reliable geometric sensing coverage in near-contact scenarios.

Rather than replacing conventional vision or tactile systems, TransTac provides a complementary sensing configuration that preserves visual transparency while enabling metric contact reconstruction and appearance consistency within a unified device.

\section{Limitation and Future Work}

Although TransTac enables integrated visual observation and tactile reconstruction, several limitations remain. First, the current system focuses primarily on geometric sensing and does not yet directly measure force or pressure distributions. Extending the sensor to support force estimation through elastomer deformation modeling or embedded sensing elements is an important direction for future work.

Second, the detection and tracking of densely distributed markers require annotated training data, and collecting labeled datasets for marker-based tactile sensing remains labor-intensive. In addition, existing depth estimation models are typically trained on datasets with limited extreme close-range or manipulation scenarios, which may lead to scale inaccuracies in near-contact depth recovery. Future work will explore improved data collection strategies and domain adaptation techniques for near-contact perception.

Third, the current prototype uses off-the-shelf stereo camera modules and illumination components. Hardware optimization, including smaller camera modules and integrated fingertip-scale designs, could further improve compactness and enable deployment on robotic grippers or dexterous hands. In addition, switching between illumination modes introduces minor latency due to camera exposure adaptation, which could be reduced through tighter hardware–software integration.

Finally, the fabrication process of ultraviolet fluorescent markers currently relies on manual or semi-automatic procedures, and the durability of the fluorescent materials under long-term use remains to be systematically evaluated. Future work will investigate automated marker deposition techniques and more robust fluorescent materials to improve manufacturability and long-term stability.

\section*{Acknowledgments}
The authors acknowledge the use of generative AI tools for language polishing and minor code prototyping assistance. All experimental design, implementation, analysis, and conclusions are solely the responsibility of the authors.
\

\bibliographystyle{IEEEtran}

\bibliography{mybib}

@article{yuan2017gelsight,
  title={Gelsight: High-resolution robot tactile sensors for estimating geometry and force},
  author={Yuan, Wenzhen and Dong, Shuang and Adelson, Edward H},
  journal={Sensors},
  volume={17},
  number={12},
  pages={2762},
  year={2017},
  publisher={MDPI}
}

@inproceedings{donlon2018gelslim,
  title={Gelslim: A high-resolution, compact, robust, and calibrated tactile-sensing finger},
  author={Donlon, Elliott and Dong, Siyuan and Liu, Melody and Li, Jianhua and Adelson, Edward and Rodriguez, Alberto},
  booktitle={2018 IEEE/RSJ International Conference on Intelligent Robots and Systems (IROS)},
  pages={1927--1934},
  year={2018},
  organization={IEEE}
}

@article{lambeta2020digit,
  title={Digit: A novel design for a low-cost compact high-resolution tactile sensor with application to in-hand manipulation},
  author={Lambeta, Mike and Chou, Po-Wei and Tian, Stephen and Yang, Brian and Maloon, Benjamin and Most, Victoria Rose and Stroud, Dave and Santos, Raymond and Byagowi, Ahmad and Kammerer, Gregg and others},
  journal={IEEE Robotics and Automation Letters},
  volume={5},
  number={3},
  pages={3838--3845},
  year={2020},
  publisher={IEEE}
}

@article{zhang2017learning,
  title={Making sense of vision and touch: Learning multimodal representations for contact-rich tasks},
  author={Lee, Michelle A and Zhu, Yuke and Zachares, Peter and Tan, Matthew and Srinivasan, Krishnan and Savarese, Silvio and Fei-Fei, Li and Garg, Animesh and Bohg, Jeannette},
  journal={IEEE Transactions on Robotics},
  volume={36},
  number={3},
  pages={582--596},
  year={2020},
  publisher={IEEE}
}

@article{sun2019fingervision,
  title={Fingervision tactile sensor design and slip detection using convolutional lstm network},
  author={Zhang, Yazhan and Kan, Zicheng and Tse, Yu Alexander and Yang, Yang and Wang, Michael Yu},
  journal={arXiv preprint arXiv:1810.02653},
  year={2018}
}

@article{lee2023tac2pose,
  title={Tac2pose: Tactile object pose estimation from the first touch},
  author={Bauza, Maria and Bronars, Antonia and Rodriguez, Alberto},
  journal={The International Journal of Robotics Research},
  volume={42},
  number={13},
  pages={1185--1209},
  year={2023},
  publisher={SAGE Publications Sage UK: London, England}
}

@inproceedings{zheng2024materobot,
  title={Materobot: Material recognition in wearable robotics for people with visual impairments},
  author={Zheng, Junwei and Zhang, Jiaming and Yang, Kailun and Peng, Kunyu and Stiefelhagen, Rainer},
  booktitle={2024 IEEE International Conference on Robotics and Automation (ICRA)},
  pages={2303--2309},
  year={2024},
  organization={IEEE}
}

@inproceedings{li2024jacquard,
  title={Jacquard v2: Refining datasets using the human in the loop data correction method},
  author={Li, Qiuhao and Yuan, Shenghai},
  booktitle={2024 IEEE International Conference on Robotics and Automation (ICRA)},
  pages={7932--7938},
  year={2024},
  organization={IEEE}
}

@inproceedings{luo2024compdvision,
  title={Compdvision: Combining near-field 3D visual and tactile sensing using a compact compound-eye imaging system},
  author={Luo, Lifan and Zhang, Boyang and Peng, Zhijie and Cheung, Yik Kin and Zhang, Guanlan and Li, Zhigang and Wang, Michael Yu and Yu, Hongyu},
  booktitle={2024 IEEE/RSJ International Conference on Intelligent Robots and Systems (IROS)},
  pages={262--268},
  year={2024},
  organization={IEEE}
}

@inproceedings{wang2022spectac,
  title={SpecTac: A visual-tactile dual-modality sensor using UV illumination},
  author={Wang, Qi and Du, Yipai and Wang, Michael Yu},
  booktitle={2022 international conference on robotics and automation (ICRA)},
  pages={10844--10850},
  year={2022},
  organization={IEEE}
}

@inproceedings{taylor2022gelslim,
  title={Gelslim 3.0: High-resolution measurement of shape, force and slip in a compact tactile-sensing finger},
  author={Taylor, Ian H and Dong, Siyuan and Rodriguez, Alberto},
  booktitle={2022 International Conference on Robotics and Automation (ICRA)},
  pages={10781--10787},
  year={2022},
  organization={IEEE}
}

@inproceedings{wen2025foundationstereo,
  title={Foundationstereo: Zero-shot stereo matching},
  author={Wen, Bowen and Trepte, Matthew and Aribido, Joseph and Kautz, Jan and Gallo, Orazio and Birchfield, Stan},
  booktitle={Proceedings of the Computer Vision and Pattern Recognition Conference},
  pages={5249--5260},
  year={2025}
}

@inproceedings{zhou2025nnwnet,
  title={nnWNet: Rethinking the Use of Transformers in Biomedical Image Segmentation and Calling for a Unified Evaluation Benchmark},
  author={Zhou, Yanfeng and Li, Lingrui and Lu, Le and Xu, Minfeng},
  booktitle={Proceedings of the Computer Vision and Pattern Recognition Conference},
  pages={20852--20862},
  year={2025}
}

@inproceedings{duan2019centernet,
  title={Centernet: Keypoint triplets for object detection},
  author={Duan, Kaiwen and Bai, Song and Xie, Lingxi and Qi, Honggang and Huang, Qingming and Tian, Qi},
  booktitle={Proceedings of the IEEE/CVF international conference on computer vision},
  pages={6569--6578},
  year={2019}
}

@inproceedings{do2022densetact,
  title={Densetact: Optical tactile sensor for dense shape reconstruction},
  author={Do, Won Kyung and Kennedy, Monroe},
  booktitle={2022 International Conference on Robotics and Automation (ICRA)},
  pages={6188--6194},
  year={2022},
  organization={IEEE}
}

@article{lin20239dtact,
  title={9dtact: A compact vision-based tactile sensor for accurate 3d shape reconstruction and generalizable 6d force estimation},
  author={Lin, Changyi and Zhang, Han and Xu, Jikai and Wu, Lei and Xu, Huazhe},
  journal={IEEE Robotics and Automation Letters},
  volume={9},
  number={2},
  pages={923--930},
  year={2023},
  publisher={IEEE}
}

@article{zhang2023tirgel,
  title={TIRgel: A visuo-tactile sensor with total internal reflection mechanism for external observation and contact detection},
  author={Zhang, Shixin and Sun, Yuhao and Shan, Jianhua and Chen, Zixi and Sun, Fuchun and Yang, Yiyong and Fang, Bin},
  journal={IEEE Robotics and Automation Letters},
  volume={8},
  number={10},
  pages={6307--6314},
  year={2023},
  publisher={IEEE}
}

@article{lu2025stereotactip,
  title={StereoTacTip: Vision-Based Tactile Sensing With Biomimetic Skin-Marker Arrangements},
  author={Lu, Chenghua and Tang, Kailuan and Hui, Xueming and Li, Haoran and Nam, Saekwang and Lepora, Nathan F},
  journal={IEEE/ASME Transactions on Mechatronics},
  year={2025},
  publisher={IEEE}
}

@article{sun2025tactile,
  title={Tactile data generation and applications based on visuo-tactile sensors: A review},
  author={Sun, Yuhao and Cheng, Ning and Zhang, Shixin and Li, Wenzhuang and Yang, Lingyue and Cui, Shaowei and Liu, Huaping and Sun, Fuchun and Zhang, Jianwei and Guo, Di and others},
  journal={Information Fusion},
  volume={121},
  pages={103162},
  year={2025},
  publisher={Elsevier}
}

@article{james2018slip,
  title={Slip detection with a biomimetic tactile sensor},
  author={James, Jasper Wollaston and Pestell, Nicholas and Lepora, Nathan F},
  journal={IEEE Robotics and Automation Letters},
  volume={3},
  number={4},
  pages={3340--3346},
  year={2018},
  publisher={IEEE}
}

@article{lin2022dtact,
  title={Dtact: A vision-based tactile sensor that measures high-resolution 3d geometry directly from darkness},
  author={Lin, Changyi and Lin, Ziqi and Wang, Shaoxiong and Xu, Huazhe},
  journal={arXiv preprint arXiv:2209.13916},
  year={2022}
}

@inproceedings{wang2025vggt,
  title={Vggt: Visual geometry grounded transformer},
  author={Wang, Jianyuan and Chen, Minghao and Karaev, Nikita and Vedaldi, Andrea and Rupprecht, Christian and Novotny, David},
  booktitle={Proceedings of the Computer Vision and Pattern Recognition Conference},
  pages={5294--5306},
  year={2025}
}

@article{ward2018tactip,
  title={The tactip family: Soft optical tactile sensors with 3d-printed biomimetic morphologies},
  author={Ward-Cherrier, Benjamin and Pestell, Nicholas and Cramphorn, Luke and Winstone, Benjamin and Giannaccini, Maria Elena and Rossiter, Jonathan and Lepora, Nathan F},
  journal={Soft robotics},
  volume={5},
  number={2},
  pages={216--227},
  year={2018},
  publisher={Mary Ann Liebert, Inc. 140 Huguenot Street, 3rd Floor New Rochelle, NY 10801 USA}
}

@inproceedings{smisek20113d,
  title={3D with Kinect},
  author={Smisek, Jan and Jancosek, Michal and Pajdla, Tomas},
  booktitle={2011 IEEE international conference on computer vision workshops (ICCV Workshops)},
  pages={1154--1160},
  year={2011},
  organization={IEEE}
}

@article{wang2025yoloe,
  title={Yoloe: Real-time seeing anything},
  author={Wang, Ao and Liu, Lihao and Chen, Hui and Lin, Zijia and Han, Jungong and Ding, Guiguang},
  journal={arXiv preprint arXiv:2503.07465},
  year={2025}
}

@inproceedings{cheng2024yolo,
  title={Yolo-world: Real-time open-vocabulary object detection},
  author={Cheng, Tianheng and Song, Lin and Ge, Yixiao and Liu, Wenyu and Wang, Xinggang and Shan, Ying},
  booktitle={Proceedings of the IEEE/CVF conference on computer vision and pattern recognition},
  pages={16901--16911},
  year={2024}
}

@inproceedings{yamaguchi2016combining,
  title={Combining finger vision and optical tactile sensing: Reducing and handling errors while cutting vegetables},
  author={Yamaguchi, Akihiko and Atkeson, Christopher G},
  booktitle={2016 IEEE-RAS 16th international conference on humanoid robots (humanoids)},
  pages={1045--1051},
  year={2016},
  organization={IEEE}
}

@article{servi2021metrological,
  title={Metrological characterization and comparison of d415, d455, l515 realsense devices in the close range},
  author={Servi, Michaela and Mussi, Elisa and Profili, Andrea and Furferi, Rocco and Volpe, Yary and Governi, Lapo and Buonamici, Francesco},
  journal={Sensors},
  volume={21},
  number={22},
  pages={7770},
  year={2021},
  publisher={MDPI}
}

@article{li2023real,
  title={Real-time and robust feature detection of continuous marker pattern for dense 3-d deformation measurement},
  author={Li, Mingxuan and Zhou, Yen Hang and Li, Tiemin and Jiang, Yao},
  journal={Measurement},
  volume={221},
  pages={113479},
  year={2023},
  publisher={Elsevier}
}

@inproceedings{geiger2010efficient,
  title={Efficient large-scale stereo matching},
  author={Geiger, Andreas and Roser, Martin and Urtasun, Raquel},
  booktitle={Asian conference on computer vision},
  pages={25--38},
  year={2010},
  organization={Springer}
}

@inproceedings{zhang2022bytetrack,
  title={Bytetrack: Multi-object tracking by associating every detection box},
  author={Zhang, Yifu and Sun, Peize and Jiang, Yi and Yu, Dongdong and Weng, Fucheng and Yuan, Zehuan and Luo, Ping and Liu, Wenyu and Wang, Xinggang},
  booktitle={European conference on computer vision},
  pages={1--21},
  year={2022},
  organization={Springer}
}

@article{tschannen2025siglip,
  title={Siglip 2: Multilingual vision-language encoders with improved semantic understanding, localization, and dense features},
  author={Tschannen, Michael and Gritsenko, Alexey and Wang, Xiao and Naeem, Muhammad Ferjad and Alabdulmohsin, Ibrahim and Parthasarathy, Nikhil and Evans, Talfan and Beyer, Lucas and Xia, Ye and Mustafa, Basil and others},
  journal={arXiv preprint arXiv:2502.14786},
  year={2025}
}

@article{oquab2023dinov2,
  title={Dinov2: Learning robust visual features without supervision},
  author={Oquab, Maxime and Darcet, Timoth{\'e}e and Moutakanni, Th{\'e}o and Vo, Huy and Szafraniec, Marc and Khalidov, Vasil and Fernandez, Pierre and Haziza, Daniel and Massa, Francisco and El-Nouby, Alaaeldin and others},
  journal={arXiv preprint arXiv:2304.07193},
  year={2023}
}

\end{document}